\documentclass[conference]{IEEEtran}
\IEEEoverridecommandlockouts
\usepackage{cite}
\usepackage{amsmath,amssymb,amsfonts}
\usepackage{graphicx}
\usepackage{textcomp}
\usepackage{xcolor}
\usepackage{hyperref}
\usepackage{cleveref}
\usepackage{algorithm}
\usepackage{algpseudocode}
\usepackage{booktabs}
\usepackage{wrapfig}
\usepackage{soul}
\usepackage{etoolbox}
\usepackage{enumitem}
\usepackage{subcaption}
\usepackage{multirow}
\usepackage{multicol}

\usepackage{float}

\newtheorem{remark}{Remark}

\def\BibTeX{{\rm B\kern-.05em{\sc i\kern-.025em b}\kern-.08em
    T\kern-.1667em\lower.7ex\hbox{E}\kern-.125emX}}
\begin{document}

\title{\LARGE \bf
Cognitively-Grounded On-Device Runtime Learning for Ground Robots in Unknown Physical Environments
}

\author{Yihao Cai$^{1}$, Yanbing Mao$^{2}$, and Christian Lebiere$^{3}$
\thanks{$^{1}$Yihao Cai is with Department of Electrical and Computer Engineering, Wayne State University, Detroit, MI, USA. {\tt\small yihao.cai@wayne.edu}}%
\thanks{$^{2}$Yanbing Mao is with the Engineering Technology Division, Wayne State University, Detroit, MI, USA. {\tt\small hm9062@wayne.edu}}
\thanks{$^{3}$Christian Lebiere is with the Department of Psychology, Carnegie Mellon University, Pittsburgh, PA, USA. {\tt\small cl@cmu.edu}}%
}

\maketitle

\begin{abstract}
This paper presents \ul{CogRun}, a framework that enables safety-critical ground robots to perform cognitively-grounded runtime learning entirely on edge-AI devices in unknown physical environments, without prior maps or perceptual knowledge. CogRun consists of three components: a Learning-Agent, a Rational-Agent, and a Coordinator. The Learning-Agent is novel in cognitive-neural learning architecture, which featurs dedicated replay buffers, cognition-driven experience sampling, and a safety-aware action blending of actor-critic reinforcement learning (RL) with instance-based learning (IBL). The Rational-Agent is a non-learning module that complements the Learning-Agent by exclusively handling safety-critical functions, while the Coordinator manages interactions between the two agents to promote safe and efficient runtime learning. CogRun's full autonomy stack (i.e., perception, learning, and control) on edge-AI devices eliminates dependence on wireless communications, enabling broader applications in challenging environments with limited or no connectivity. Experiments on a quadruped robot in real-world wild forests and on an off-road autonomous vehicle in a simulated wild forest demonstrate that CogRun enables safe and efficient runtime learning, allowing robots to safely and continuously interact with the physical world for enhancing task performance in complex, unknown environments.

\end{abstract}


\section{Introduction}
AI is increasingly being integrated into ground robots, such as legged robots and autonomous vehicles, to achieve human-level autonomy. However, these advances have primarily focused on human-centered and structured environments. Many emerging applications, including navigation and exploration in post-disaster areas and wild forests, require robots to operate in unstructured, dynamic, and unpredictable physical environments. Achieving human-level autonomy in these environments faces the following fundamental challenges. 

\textbf{Challenge 1: Non-Stationary and Unknown Physical Environments.}
Existing AI-embodied robots predominantly follow a train-then-deploy paradigm, where the intelligence is learned offline and deployed as a fixed capability. This paradigm inevitably suffers from domain and sim-to-real gaps, resulting in degraded performance when operating in environments beyond the training distribution. Considerable progress has been made in reducing these gaps through simulation, domain adaptation, and transfer learning \cite{nagabandi2018learning,cloudedge,imai2022vision,yang2022safe,tan2018sim}. However, domain and sim-to-real gaps remain unavoidable in many real-world scenarios, such as wild forests and post-natural disaster environments, which are are non-stationary, unknown, and impossible to fully characterize before deployment. 


\textbf{Challenge 2: Time-Critical Missions in Communication-Limited Environments.}
Current robot learning frameworks typically rely on cloud–edge computing architectures \cite{runtimemao,realdrl,cloudedge,Liu_2019}, which assume reliable wireless connectivity and access to remote computational resources. However, ground robots often operate in wild, remote, and chaotic physical environments where wireless communication is unreliable, intermittent, or entirely unavailable, fundamentally limiting the practicality of cloud-assisted learning. Furthermore, the inherent communication latency of cloud–edge architectures makes them unsuitable for safety- and time-critical robotic missions that require real-time perception, decision-making, and adaptation.

\textbf{Challenge 3: Experience Replay for Efficient Robot Learning.}
Robots naturally generate experiences that are temporally correlated with their operating state. However, most existing experience replay strategies do not account for the relevance of replayed experiences to the robot's current state, leading to inefficient usage of valuable real-world data. This issue is exacerbated in complex, previously unseen, and non-stationary environments, where outdated or irrelevant experiences can significantly degrade roboot learning efficiency. On-policy paradigms inherently mitigate this issue but badly suffer from poor sample efficiency \cite{margolis2024rapid,hoeller2024anymal,radosavovic2024real}, making them impractical for long-lasting runtime learning on physical robots. 

\textbf{Approach:} To address \textbf{Challenges 1--3}, we propose \textbf{CogRun}, a cognitively-grounded on-device runtime learning framework for ground robots. Specifically,
\begin{itemize}
\item \textbf{CogRun} enables robots to perform safe runtime learning in complex physical environments, allowing continuous adaptation to non-stationary and previously unknown operating conditions for addressing \textbf{Challenge 1}.
\item \textbf{CogRun} establishes a complete runtime learning stack on edge-AI devices, eliminating dependence on cloud--edge communication while enabling lightweight, low-latency, and temporally consistent onboard execution for addressing \textbf{Challenge 2}.
\item \textbf{CogRun} introduces a novel cognitive--neural learning architecture that selectively prioritizes experiences most relevant to the robot's current runtime state, enabling experience-efficient runtime learning from scarce real-world interactions for addressing \textbf{Challenge 3}.
\end{itemize}

\noindent Finally, we implement \textbf{CogRun} in ROS2 on an embedded NVIDIA Jetson AGX Orin edge-AI platform and validate it through real-world experiments on a quadruped robot operating in complex, unknown forest environments, as well as on an off-road autonomous vehicle in a simulated wild forest. Experimental results on two representative tasks, runtime learning for navigation and end-to-end vision-to-action learning, demonstrate that \textbf{CogRun} significantly enhance safety assurance, experience efficiency, and task performance of runtime learning.

\section{Related Work}\label{Related}



\textbf{Runtime Learning in Physical World:} It has recently been recognized that runtime learning is essential for addressing \textbf{Challenge 1} \cite{key1,key2,dunlap2025demonstrating}. Notably, many robots are safety-critical, where even a slight safety violation can result in catastrophic consequences, demanding rigorous safety assurance for runtime learning framework. Runtime-assurance reinforcement learning (RL) offers a pragmatic alternative for enabling runtime learning in safety-critical autonomous systems. Existing frameworks include neural Simplex \cite{phan2020neural}, runtime assurance \cite{brat2023runtime,sifakis2023trustworthy,chen2022runtime}, and model predictive shielding \cite{bastani2021safe,banerjee2024dynamic}. These approaches treat RL as a high-performance yet black-box module operating in parallel with a high-assurance module, which is designed to back up system safety by tolerating RL faults. Building on this logic, a few runtime learning frameworks have been proposed \cite{runtimemao,realdrl}. However, they rely on cloud-edge communication, and thus face \textbf{Challenge 2}, which limits their application domains. 

\textbf{Context-Aware Experience Replay:} Experience replay is widely adopted to improve sample efficiency, with most methods relying on general-purpose RL strategies such as uniform sampling, prioritized experience replay (PER), and mixed replay 
\cite{rana2021bayesian,li2022equipping,cheng2019control,schaul2016prioritizedexperiencereplay,Phydrl1,YU2024124017}. 
Recent work has explored replay under non-stationary environments 
\cite{duan2025sampleefficientexperiencereplay}, state-aware relevance 
\cite{Sun2020AttentiveER}, and policy-proximity relevance 
\cite{novati2019rememberforgetexperiencereplay}. However, these methods measure relevance mainly through a single geometric or policy-based signal, neglecting the safety-critical nature and high interaction cost of physical robots, and lack a principled memory mechanism that captures experience relevance over time, making them insufficient for safe and efficient runtime learning on physical robots. Consequently, they cannot effectively address \textbf{Challenge 3}.

\newlength{\customwidth}      
\newlength{\fighspace}        
\setlength{\customwidth}{0.0\textwidth}
\setlength{\fighspace}{-.000cm}
\begin{figure*}[!http]
\centerline{\includegraphics[width=0.999\textwidth]{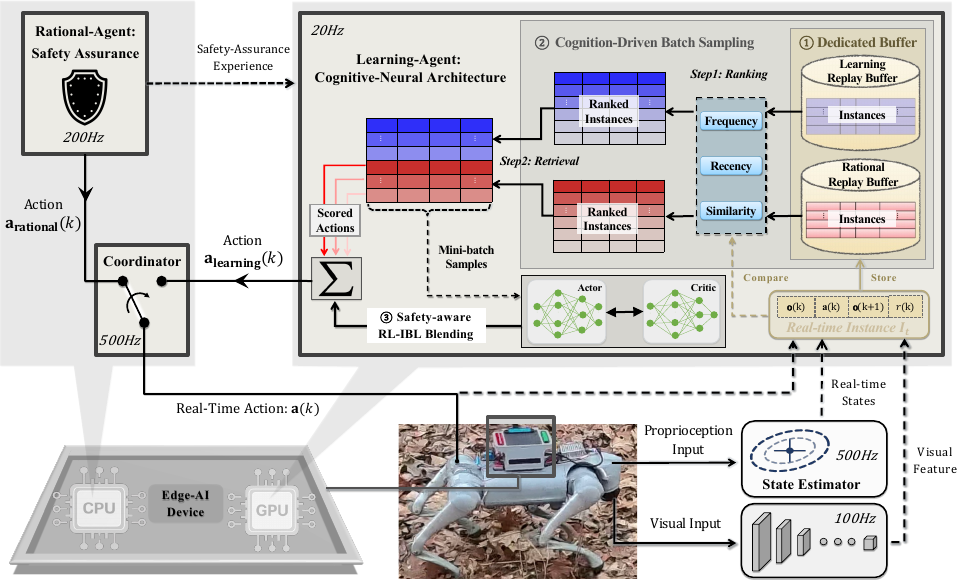}}
\caption{The proposed CogRun framework deployed on Edge-AI device for on-device robot runtime learning.} 
\vspace{0.15cm}
On an Edge-AI device, the robot states (500 Hz) and visual features (100 Hz) are provided by the state estimator and feature encoder, respectively. The \textbf{Rational-Agent} generates safety-assured actions (200 Hz) and stores safety-assurance experiences in the Rational Replay Buffer, while the \textbf{Learning-Agent} interacts with the physical environment at 20 Hz and stores trial-and-error experiences in the Learning Replay Buffer. Safe and efficient runtime learning is achieved through a cognitive-neural architecture with (1) dedicated replay buffers, (2) cognition-driven batch sampling, and (3) safety-aware RL-IBL blending. The \textbf{Coordinator} managers the two agents based on the real-time robot states. The architecture adopts a decoupled CPU-GPU execution: the \textbf{Rational-Agent} and \textbf{Coordinator} run on the \textbf{CPU}; the \textbf{Learning-Agent} runs on the \textbf{GPU}.

\label{learningframework}
\vspace{-0.35cm}
\end{figure*}
\section{Methodology: CogRun Framework}
As shown in Fig.~\ref{learningframework}, \textbf{CogRun} builds on advances in cognitive science and reinforcement learning, and consists of three tightly coupled components:
\begin{itemize}
\item \textbf{Learning-Agent}, which builds upon an actor--critic RL framework and introduces a cognitively grounded neural learning architecture for runtime learning. Its key innovations include cognition-driven batch sampling and real-time action blending between reinforcement learning and instance-based learning.
\item \textbf{Rational-Agent}, which complements the Learning-Agent in assuring robot safety, as it is a non-learning-based design focused \ul{exclusively on safety-critical functions}.
\item \textbf{Coordinator}, which orchestrates the interaction between the Learning-Agent and the Rational-Agent to enable safe and efficient runtime learning in the physical world. Specifically, when the Learning-Agent generates an unsafe action during runtime learning, the Rational-Agent intervenes to guarantee robot safety while sharing its safety-assurance experience with the Learning-Agent, thereby reducing risky trial-and-error exploration.
\end{itemize}
We next describe the detailed design of these three interactive components and their operation practice on edge-AI devices.

\subsection{Coordinator Component}\label{coordinator}
We consider the ground robots as safety-critical systems requiring their system states to always remain within:
\begin{equation}
\text{Safety Set:}~~{\mathbb{S}} \triangleq \left\{ {\left. \mathbf{s} \in {\mathbb{R}^n} \right|-\mathbf{c} \le \mathbf{C} \cdot \mathbf{s}  \le \mathbf{c}} \right\} \label{aset2} 
\end{equation}
where $\mathbf{s} \in \mathbb{R}^{n}$ denotes system state vector. $\mathbf{C} \in \mathbb{R}^{p \times n}$ and $\mathbf{c} \in \mathbb{R}^p$ are pre-determined to describe $p \in \mathbb{N}$ safety conditions. Examples include lane tracking \cite{lee2025lane}, velocity regulation \cite{amac,rajamani2011vehicle} and collision avoidance \cite{he2024interactive}, amongst many others. 

The Coordinator monitors the robot's real-time states and manages interactions between the Learning-Agent and Rational-Agent according to conditions for defining safety set $\mathbb{S}$ \eqref{aset2}. Referring to Fig. \ref{learningframework}, the terminal real-time action $\mathbf{a}(t)$ applied to the real robot is as follows:
\begin{align}
\mathbf{a}(t) \leftarrow \begin{cases}
		\mathbf{a}_{\text{learning}}(t), &\text{if}~\mathbf{s}(t) \in \mathbb{S},\\ 
        \mathbf{a}_{\text{rational}}(t),      &\text{if}~\mathbf{s}(t) \notin \mathbb{S},
	\end{cases} \label{terminalaction}
\end{align}
where $\mathbf{a}_{\text{learning}}(t)$ and $\mathbf{a}_{\text{rational}}(t)$ are computed by the Learning-Agent and Rational-Agent, respectively, in parallel. The Rational-Agent provides verifiable safety assurance and, when activated, temporarily takes over control to maintain  robot safety while sharing safety-critical experiences for the Learning-Agent in learning safe behaviors. Once the system state returns to the interior of the safety set, control is automatically handed back to the Learning-Agent, and safety-assured experience collection is suspended.

\subsection{Rational-Agent Component}\label{rational}
The Rational-Agent module is a non-learning-based design that focuses exclusively on \ul{safety-critical} functions. It acts as a safety fallback, intervening whenever the Learning-Agent risks a safety violation during runtime learning. Built upon well-established methods with verifiable safety guarantees, the Rational-Agent enables safe runtime learning in non-stationary physical environments, including those with unknown or uncertain physical conditions. Although such methods can rigorously ensure system safety, they are generally conservative and unable to achieve high task performance, which is attainable by learning-based policies. Depending on the task objective, the Rational-Agent can be instantiated using different safety-assured methods, such as physics-model-based approaches for control policy learning \cite{phan2020neural,brat2023runtime,sifakis2023trustworthy,chen2022runtime,bastani2021safe,banerjee2024dynamic,realdrl}, or Interactive-Far \cite{he2024interactive} combined with a physics-model-based safety module for end-to-end vision-to-action learning.

\subsection{Learning-Agent Component}\label{learning}
Showing in Fig. \ref{learningframework}, Learning-Agent features a cognitive-neural architecture for runtime learning, with innovations in: 1) dedicated replay buffers, 2) cognition-driven batch sampling, and 3) a safety-aware blending of RL and IBL. 

\subsubsection{\textbf{Dedicated Replay Buffers}}
As introduced in \cref{coordinator}, the Rational-Agent shares its safety-assurance experiences with the Learning-Agent. Two dedicated replay buffers, $\mathcal{B}_{\mathrm{L}}$ and $\mathcal{B}_{\mathrm{R}}$, are proposed to store transitions $\tau_t^{\mathrm{L}}$ and $\tau_t^{\mathrm{R}}$ of Learning-Agent and Rational-Agent, respectively:
\[
\left\{
\begin{aligned}
\mathcal{B}_{\mathrm{L}}
&:
\left\{
\tau_t^{\mathrm{L}}
=
(\mathbf{o}_t,\mathbf{a}_t^{\mathrm{L}},
\mathbf{o}_{t+1},r_t)
\right\}_{t=1}^{|\mathcal{B}_{\mathrm{L}}|},
~~~
\mathbf{s}_t\in\mathbb{S}
\\
\mathcal{B}_{\mathrm{R}}
&:
\left\{
\tau_t^{\mathrm{R}}
=
(\mathbf{o}_t,\mathbf{a}_t^{\mathrm{R}},
\mathbf{o}_{t+1},r_t)
\right\}_{t=1}^{|\mathcal{B}_{\mathrm{R}}|},
~~\mathbf{s}_t\notin\mathbb{S}
\end{aligned}
\right.
\]
where $|\cdot|$ denotes the buffer size. $\mathbf{o}_t=[\mathbf{s}_t,\mathbf{z}_t]\in\mathbb{R}^n$ is the stacked vector of proprioceptive states $\mathbf{s}_t$ and encoded visual features $\mathbf{z}_t$ at time $t$. $\mathbf{a}_t^{\mathrm{L}}=\mathbf{a}_{\text{learning}}(t)$ and $\mathbf{a}_t^{\mathrm{R}}=\mathbf{a}_{\text{rational}}(t)$ are the
actions from the Learning-Agent and Rational-Agent, respectively. $r_t = r(\mathbf{o}_t,\mathbf{a}_t,\mathbf{o}_{t+1})$
denotes the stepwise reward, with $r:\mathcal{O}\times\mathcal{A}\times\mathcal{O}\rightarrow\mathbb{R}$
as the reward function.

The $\mathcal{B}_{\mathrm{L}}$ stores task-oriented experiences collected through the Learning-Agent's trial-and-error exploration, while the $\mathcal{B}_{\mathrm{R}}$ stores safety-assured experiences generated by the Rational-Agent during safety interventions. As illustrated in Fig. \ref{learningframework}, each mini-batch $\mathcal{M}$ for runtime learning is constructed by sampling transitions from both replay buffers:
\begin{align}
|\mathcal{M}|
=
|\mathcal{M}_{\mathrm{L}}|
+
|\mathcal{M}_{\mathrm{R}}|,
\label{bsma}
\end{align}
where $\mathcal{M}_{\mathrm{L}}\subset\mathcal{B}_{\mathrm{L}}$ and $\mathcal{M}_{\mathrm{R}}\subset\mathcal{B}_{\mathrm{R}}$ are the sub-batches sampled from the Learning Replay Buffer and Rational Replay Buffer, respectively. By incorporating assuring-safety experiences into the mini-batch, the Learning-Agent can continuously learn safe strategies from the Rational-Agent, improving sample efficiency and reducing risky trial-and-error exploration.


\subsubsection{\textbf{Cognition-Driven Batch Sampling}}\label{insightssuma}
Existing experience replay strategies in general-purpose RL either rarely consider the relationship between collected experiences and the robot's current state, or lack a joint metric for capturing it. In contrast, cognitive science implies human memory retrieval is guided by several fundamental principles, enabling context-dependent recall of the most relevant experiences:

\begin{itemize} \label{threekeycogprinciple}
\vspace{-0.05cm}

\item \textit{Similarity}:
Human memory retrieval is guided by both surface-level and structural similarity to the current context, with structural similarity playing a dominant role in relational reasoning \cite{anderson1997act,anderson2004integrated,lebiere2013functional,lebiere2012functional,thomson2015general}.
\item \textit{Frequency}:
The human brain naturally prioritizes familiar and frequently encountered experiences, improving retrieval efficiency through repeated exposure \cite{popov2020frequency,berglund2019word}.
\item \textit{Recency}:
Recently acquired experiences are more readily recalled, facilitating efficient retrieval and rapid adaptation in dynamic environments \cite{sederberg2008context,holm2023reliable}.
\end{itemize}
The insights from human cognition motivate our cognition-driven batch sampling mechanism for experience-efficient runtime learning, which consists of the following two steps. 

\ul{Step 1: Ranking:}
Each $i$-th stored transition $\tau_i$ is regarded as an experience instance, denoted by $\mathbf{I}_i$. In particular, $\mathbf{I}_t$ denotes the real-time experience instance collected at time $t$, characterizing the robot's current operating state:
\begin{align}
\mathbf{I}_t \triangleq
\begin{cases}
\tau_t^{\mathrm L}
=
(\mathbf{o}_t,\mathbf{a}_t^{\mathrm L},
\mathbf{o}_{t+1},
r_t),
& \mathbf{s}_t\in\mathbb S,\\[1mm]
\tau_t^{\mathrm R}
=
(\mathbf{o}_t,\mathbf{a}_t^{\mathrm R}
\mathbf{o}_{t+1},
r_t),
& \mathbf{s}_t\notin\mathbb S
\end{cases}
\label{insdef}
\end{align}
For each experience instance $\mathbf{I}_i \in \mathcal{B}_{\sigma}$,
$i=1,2,\ldots,|\mathcal{B}_{\sigma}|$, where $\sigma \in \{\mathrm{L},\mathrm{R}\}$, its ranking score with respect to the real-time experience instance $\mathbf{I}_t$ is computed as:
\begin{equation}
\text{rank}(\mathbf{I}_{i}, \mathbf{I}_t) =  \mathcal{S}(\mathbf{I}_{i}, \mathbf{I}_t) + \mathcal{F}(\mathbf{I}_{i}) + \mathcal{R}(\mathbf{I}_{i}) 
\label{rank}
\end{equation}
where $\text{rank}(\cdot,\cdot)\in\mathbb{R}$ is the ranking score of an experience instance relative to the current experience instance $\mathbf{I}_t$. The functions $\mathcal{S}(\cdot,\cdot)\in\mathbb{R}$, $\mathcal{F}(\cdot)\in\mathbb{R}$, and $\mathcal{R}(\cdot)\in\mathbb{R}$ measure similarity, frequency, and recency, respectively. 

\ul{Step 2: Retrieval:}
Each sample batch includes two sub-batches independently sampled from two dedicated buffers $\mathcal B_{\mathrm L}$ and $\mathcal B_{\mathrm R}$.
For each buffer $\mathcal B_\sigma$, where $\sigma\in\{\mathrm L,\mathrm R\}$,
we let $K_\sigma$ denote the desired sub-batch size. The sampled sub-batch is as the
top-$K_\sigma$ ranked experiences in $\mathcal B_\sigma$, according to their associated batch scores:
\begin{align}
\mathbf{S}_\sigma
\triangleq
\left[
\operatorname{rank}(\mathbf{I}_1,\mathbf{I}_t);\,
\operatorname{rank}(\mathbf{I}_2,\mathbf{I}_t);\,
\ldots;\,
\operatorname{rank}(\mathbf{I}_{K_\sigma},\mathbf{I}_t)
\right],
\label{barchscore}
\end{align}
where the score elements are decreasingly ordered.

\ul{Measures of Frequency, Recency, and Similarity:} We note that a core building block of our cognition-driven batch sampling is the measure functions of similarity, frequency, and recency. In cognitive science, the proposed time dynamics of human declarative memory is shown to simultaneously capture frequency and recency factors in memory retrieval \cite{anderson2004integrated,stanley2016comparing}. Building on this, we propose a variant that can serve as concurrent frequency and recency measure function for actor-critic RL with a continuous experience space. This mechanism is formally described in Algorithm \ref{fralg}.  The similarity score $\mathcal{S}(\mathbf{I}_{i}, \mathbf{I}_{t})$ can be instantiated by various metrics. In this work, we use
$\mathcal{S}(\mathbf{I}_{i}, \mathbf{I}_{t})=(1+\|\mathbf{I}_{i}-\mathbf{I}_{t}\|_2)^{-1}$,
where $\|\cdot\|_2$ denotes the Euclidean norm.

\begin{algorithm}[http]
\caption{Concurrent Frequency and Recency Measures}
\label{fralg}
\small
\textbf{Input:} Replay buffer $\mathcal{B}_{\sigma}$,
$\sigma\in\{\mathrm{L},\mathrm{R}\}$;
family number $\mathsf{F}\in\mathbb{N}$; family criterion $\mathsf{c}>0$; memory decay factor $\lambda>0$.
\begin{algorithmic}[1]
\State Infuse each experience instance  according to:
\begin{align}
\mathcal{I}(\mathbf{I}_i) = (\left\| \mathbf{o}_i\right\|_2 + \left\| \mathbf{a}_i\right\|_2) \cdot e^{r_{i}}, i \in \{1, 2, \ldots, |\mathcal{B}_{\sigma}| \} \label{exinfuse}
\end{align}
where $\mathbf{o}_i$, $\mathbf{a}_i$, and $\mathcal{R}_{i}$ denote the state tensor, action tensor, and reward associated with experience instance $\mathbf{I}_i$, respectively; \label{algini}
\State Identify the family ID for each infused experience instance:
\begin{align}
\mathcal{D}(\mathbf{I}_i) = \begin{cases}
		\left\lfloor \mathcal{I}(\mathbf{I}_i) / \mathsf{c}   \right\rfloor, &\text{if}~\mathcal{I}(\mathbf{I}_i) < \mathsf{F} \cdot \mathsf{c}\\ 
        \mathsf{F},      &\text{otherwise}
	\end{cases}, \label{familyid}
\end{align}
where $\left\lfloor \cdot   \right\rfloor$ denotes the floor function; \label{selfactionc2}
\State Compute the number of members within each family ID:
\begin{align}
\mathsf{N}_{j} = \mathsf{sum}(\mathcal{D}(\mathbf{I}) == j), ~~~j \in \{0, 1, \ldots, \mathsf{F}\} \label{familyidnum}
\end{align}
where `==' denotes the equality comparison operator; 
\State Compute concurrent frequency and recency measures:
\begin{align}
\mathcal{F}(\mathbf{I}_{i}) + \mathcal{R}(\mathbf{I}_{i}) = \log \left(\sum_{k=1}^{\mathsf{N}_{j}}\Delta^{-\lambda}_{j,k} \right)~\mathsf{with}~\mathcal{D}(\mathbf{I}_i) = j,\label{fr}
\end{align}
where $j \in \{0, 1, \ldots, \mathsf{F}\}$ and $\lambda > 0$ is the memory decay factor, and $\Delta_{j,k}$ is the time elapsed since the $k$-th activation of family with ID $j$, which can be readily read from the stored order in replay buffer $\mathcal{B}_{\sigma}$.
\label{selfactionc3}
\end{algorithmic}
\end{algorithm}

\begin{remark}
The model \eqref{fr} can concurrently capture frequency and recency in experience retrieval. Specifically, for the experience instances within same family $j$, a larger $\mathsf{N}_{j}$ indicates a greater number of instances belonging to the family $j$, reflecting a higher frequency of occurrence. Such high-frequency instances consequently receive a higher ranking value in Eq. \eqref{rank}. A smaller time elapsed $\Delta_{j,k}$ indicates more recently occurring experience instances. Accounting for the memory decay factor $\lambda > 0$, a smaller $\Delta_{j,k}$ also results in a higher ranking value. 
\end{remark}

\subsubsection{\textbf{Safety-Aware RL--IBL Blending}}
Instance-based learning (IBL) theory \cite{gonzalez2003instance,gonzalez2013boundaries} 
employs a blending mechanism \cite{langley2013central,lebiere1999blending} to model how humans integrate retrieved instances into contextually appropriate solutions. This insight motivates us to develop the safety-aware RL--IBL blending mechanism, which blends the actions of rational experience instances in the batch samples to derive the experience action:
\begin{equation}
\mathbf{a}_{\text{experience}} =  \sum^{|\mathcal{M}_{\mathrm R}|}_{i=1} [\text{softmax}(\textbf{S}_{\mathrm R})]_{i} \cdot \mathbf{a}_i, \label{expaction} 
\end{equation}
where $\mathbf{a}_i$ denote the actions associated with experience instance $\mathbf{I}_i$ (see \cref{insdef}) included in the batch samples, and the score batch $\textbf{S}_{\mathrm R}$ is defined in \cref{barchscore}. 

The Learning-Agent now has two types of actions at its disposal: the experience action $\mathbf{a}_{\text{experience}}$ in \cref{expaction} and the action output $\mathbf{a}_{\text{actor}}$ from the actor network. With the introduction of a state-depdent blending parameter $\eta(\mathbf{s}) \in [0,1]$, these two actions are further blended to produce the final learning action $\mathbf{a}_{\text{learning}}$ that the Learning-Agent can apply to the robot in physical world:
\begin{align}
\mathbf{a}_{\text{learning}} = (1-\eta(\mathbf{s}))  \cdot  \mathbf{a}_{\text{actor}}   + \eta(\mathbf{s})  \cdot \mathbf{a}_{\text{experience}},\label{learninga}
\end{align}
The experience action $\mathbf{a}_{\text{experience}}
$ \eqref{expaction} is generated by blending actions exclusively from rational experience instances. Intuitively, if the experience action can uphold safety assurance, it can further strengthen the safety of the terminal learning action \eqref{learninga} through controlling $\eta(\mathbf{s})
$. Motivated by this, we design the $\eta(\mathbf{s})
$ to be safety-aware, formulated as: 
\begin{align}
\eta  = \min\left\{\mathbf{s}^{\top}  \cdot  \mathbf{P} \cdot  \mathbf{s}, ~1\right\} \in [0,1],\label{learningaab}
\end{align}
where $\mathbf{P}$ is computed according to the following convex optimization (see toolbox solvers \cite{boyd1994linear,grant2009cvx}):
Let $\bar{\mathbf{C}} \triangleq \mathbf{C}\cdot\textbf{diag}^{-1}\{\mathbf{c}\}$. The matrix $\mathbf{P}$ is then computed as:
\begin{align}
\mathbf{P} = \arg\min_{\mathbf{Q}\succ 0}\{\log(\det(\mathbf{Q}))\}, ~\text{s.t.}~\mathbf{I}_p - \bar{\mathbf{C}}\mathbf{Q}^{-1}\bar{\mathbf{C}}^\top \succ 0,
\label{ssind2}
\end{align}
with $\mathbf{C}$ and $\mathbf{c}$ given in Eq. \eqref{aset2} for defining the safety set, and $\mathbf{s}$ in Eq. \eqref{learninga} denotes real-time system state vector.

\subsection{Operation Practice on Edge-AI Devices}\label{edge}
Edge-AI devices support lightweight, locally executable models, enabling direct on-device and real-time interactions, which preserves temporal consistency and low latency. In this work, we adopt a representative edge-AI device, the NVIDIA Jetson, which comprises an ARM-based CPU and a GPU. We next outline the key operation practices.  

\ul{Efficiency and Scalability:} \textbf{CogRun} deploys its Rational-Agent on an ARM-based CPU, leveraging its superior performance-per-watt ratio and thermal efficiency, both of which are crucial for mobile autonomous systems and long-duration on-device runtime learning. This design enables energy-efficient execution and scalable, robust decision-making for safety-critical edge deployment.

\ul{Decoupled Multi-Rate Execution:} \textbf{CogRun} leverages the multi-rate architecture of edge-AI devices. Specifically, the Coordinator and Rational-Agents are implemented in C/C++ on the CPU to ensure high-frequency operation (200--1000 Hz) for safe control, while the Learning-Agent runs at lower frequencies (10--20 Hz) on the GPU. By decoupling learning from safe control, the system ensures stable, safe interaction with time-critical physical environments while preserving real-time execution. This design further improves runtime learning stability without compromising real-time system safety.

\section{Experiment}
This section presents experiments on a quadruped robot and an off-road autonomous vehicle.

\subsection{Quadruped Robot in Real Unknown Wild Forests}
To demonstrate \textbf{CogRun}'s generalizability across functions, we evaluate it on two distinct tasks: navigation and end-to-end vision-to-action navigation,  coupled with comparison studies. The specifications of the Edge-AI hardware and RL network architecture are summarized in Table \ref{table:specs}.

\begin{table}[t]
\centering
\footnotesize
\renewcommand{\arraystretch}{0.98}
\begin{tabular}{@{}c|c|c@{}}
\toprule
\textbf{Category} & \textbf{Component} & \textbf{Configuration} \\
\midrule
\multirow{7}{*}{\textbf{Edge-AI}} & AI Performance           & Up to 275 TOPS \\
                          & CPU                      & 12-core Cortex-A78AE, 2.2\,GHz \\
                          & GPU                      & 2048 CUDA cores, 64 Tensor Cores \\
                          & Memory                   & 64\,GB LPDDR5 (256-bit) \\
                          & Power                    & 30\,Watts \\
                          & Software                 & JetPack 6.2, ROS\,2 Humble \\
                          & Kernel                   & PREEMPT\_RT real-time kernel \\
\midrule
\multirow{8}{*}{\textbf{RL}}       & Algorithm                & DDPG (Actor-Critic) \\
                          & Policy Network           & MLP (hidden [512, 256, 128]) \\
                          & Critic Network           & MLP (hidden [512, 256, 128]) \\
                          & Activation               & ReLU \\
                          & Optimizer                & Adam \\
                          & Batch Size               & 256 \\
                          & Buffer Capacity          & $1\times10^{6}$ \\
                          & Discount Factor $\gamma$ & 0.99 \\
\bottomrule
\end{tabular}
\caption{Edge-AI Hardware and RL configurations.} \label{table:specs}
\vspace{-0.3cm}
\end{table}

\subsubsection{\textbf{Runtime Learning for Navigation}} 
\begin{figure}[H]
\centering
\includegraphics[width=0.49\textwidth]{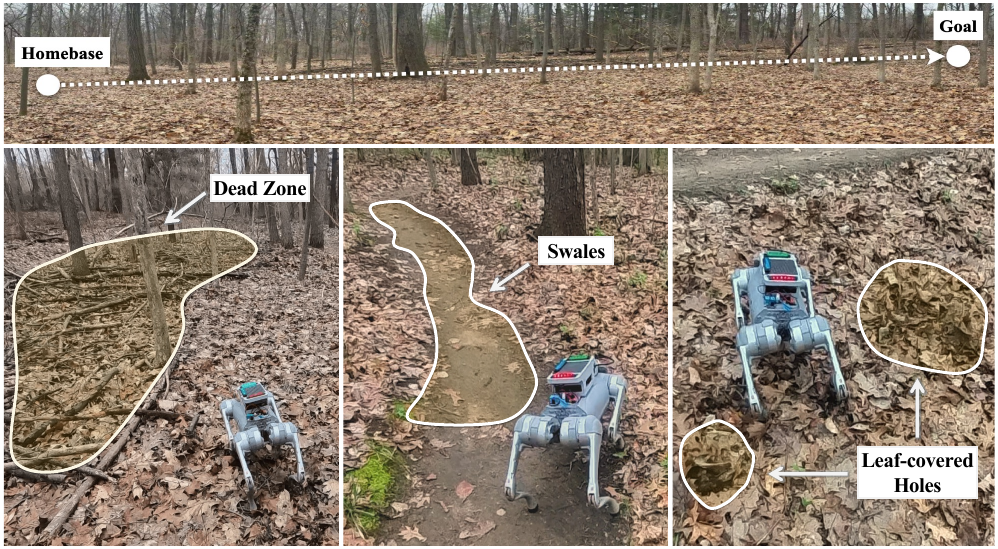}
\caption{The unknown physical environment for the quadruped robot's runtime navigation learning task (30 meters long), featuring \textit{dead zones}, \textit{swales}, and \textit{leaf-covered holes}.}
\label{env1}
\vspace{-0.3cm}
\end{figure}

The unknown wild forest is shown in Fig. \ref{env1}, featuring unstructured terrains, randomly distributed trees, dead zones, swales, and leaf-covered holes. The robot must safely navigate from the home base to a goal approximately 30 meters away. The safety set is defined as $\mathbb{S}=\{\mathbf{s}\mid d_{min}\ge1.2~\mathrm{meter},~|h-0.3|\le0.15~\mathrm{meter}\}$, where $d_{min}$ is obstacle distance and $h$ is the robot height. 

For Learning-Agent, its observation is constructed as $\mathbf{o}_t=[\mathbf{s}_t,\mathbf{z}_t]$, where $\mathbf{s}_t\in\mathbb{R}^{12}$ is the estimated Center-of-mass (CoM) state and $\mathbf{z}_t\in\mathbb{R}^{24}$ denotes the visual feature. Its action is a twist command $\mathbf{a}=[v_x^{cmd},v_y^{cmd},\omega_z^{cmd}]$, which will be sent to low-level controller for motion tracking. The reward functions design primarily considers several key aspects: 
\begin{itemize}
\item \ul{Body Stability:} Account for both stability and partial safety considerations:
\begin{equation}
R_{bs}
=
-
(\mathbf{s}_t-\mathbf{\bar{s}}_t)^{\top}
\mathbf{W}
(\mathbf{s}-\mathbf{\bar{s}}_t), \label{sta}
\end{equation}
where $\mathbf{\bar{s}}_t$ is the desired state and $\mathbf{W}$ is a weighting matrix. 
\item \ul{Path Safety:} Encourage the robot to seek safer paths by maximizing the minimum distance with obstacles:  
\begin{align}
\mathcal{R}_{ps} = - \frac{1}{d_{\min}+\epsilon}, \label{pathsafety}
\end{align}
where $d_{\min}$ denotes the minimum distance between the robot and its surrounding obstacles. $\epsilon$ is a small constant to avoid numerical singularities.
\item \ul{Goal Reaching:} encourages shorter paths to the goal:
\begin{align}
\mathcal{R}_{gr} = - ||p_{\text{position}} - p_{\text{goal}}||^{2}_{2}, \label{goaldistance} 
\end{align}
where $p_{\text{position}}$ and $p_{\text{goal}}$ denote the robot's current position and the preset goal position, respectively. 
\item \ul{Rational-Agent Activation:} Penalize Rational-Agent interventions for Learning-Agent exploration:
\begin{align}
\mathcal{R}_{RA} = \begin{cases}
		-1, &\text{if Rational-Agent is activated},\\ 
        0,      &\text{otherwise}.
	\end{cases} \label{activation} 
\end{align}
\item \ul{Travel Velocity:} Encourage efficient forward motion by penalizing slow forward and lateral velocities:
\begin{align}
\mathcal{R}_{vel}
=-
\left(
\frac{\omega_1}{|v_x|+\epsilon}
+\omega_2 \cdot v_y^2
\right),
\label{velocity}
\end{align}
where $v_x$ and $v_y$ are the forward and lateral CoM velocities, respectively, $\omega_1$ and $\omega_2$ are weighting coefficients.
\end{itemize}
The composite reward function for guiding the Learning-Agent in acquiring a safe and high-performance navigation policy through runtime learning is formulated by combining the aforementioned reward components \eqref{sta}--\eqref{velocity}: 
\begin{align}
r = \xi_1 \cdot \mathcal{R}_{bs} + \xi_2 \cdot \mathcal{R}_{ps} + \xi_3 \cdot \mathcal{R}_{gr} +  \xi_4 \cdot \mathcal{R}_{RA} + \xi_5 \cdot \mathcal{R}_{vel} \nonumber
\end{align}
where $\xi_1 = 0.5, \xi_2 = 0.05, \xi_3 = 0.02, \xi_4 = 0.5$, and $\xi_5 = 0.002$ are the corresponding weighting coefficients.

As for Rational-Agent, we adopts Interactive-FAR~\cite{he2024interactive}, which generates verifiable collision-free motions. 

We first present a demonstration video available at \href{https://www.dropbox.com/scl/fi/rjak668semgsyzpz1oayk/learning-process2.0.mp4?rlkey=6wz152ns4aotten5mliwgb264&st=vvxabhlf&dl=0}{\color{blue} [\textbf{CogRun}-Nav link (2$\times$)]}, which shows that 1) \textbf{CogRun} can strictly assure robot safety (i.e., collision and fall-down avoidance) throughout the runtime learning process, and 2) the robot continuously improves its navigation policy through interaction with the unknown environment, significantly reducing the time to goal and achieving stable and efficient navigation within only tens of learning episodes.

\newcommand{\subfigwidth}{0.232\textwidth}
\begin{figure}[!ht]
\vspace{-0.2cm}
\centering
\begin{subfigure}{\subfigwidth}
    \includegraphics[width=\linewidth]{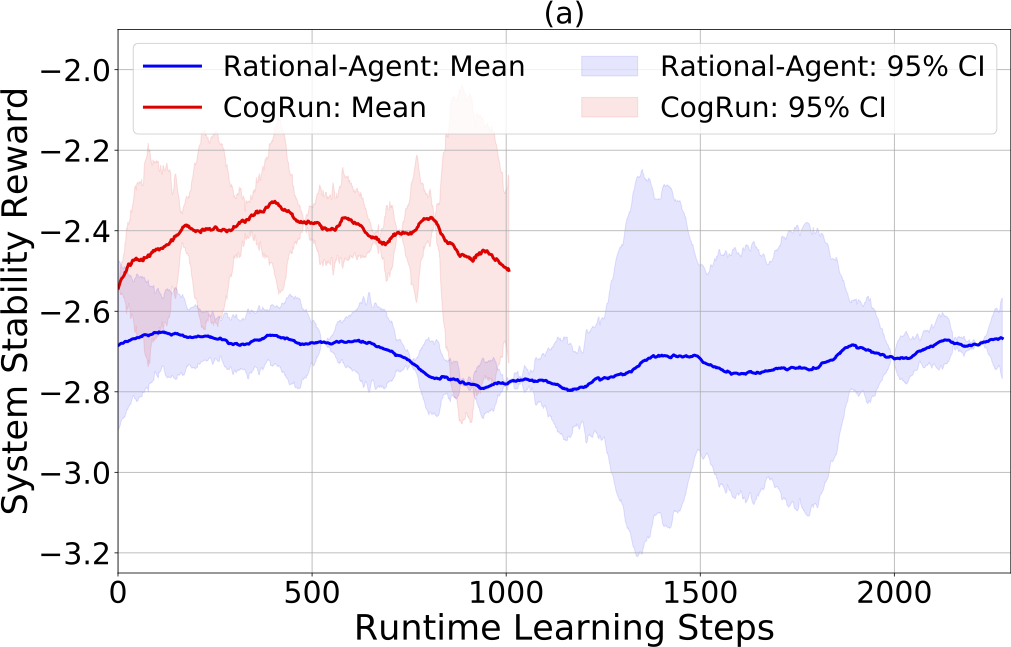}
\end{subfigure}
\hfill
\begin{subfigure}{\subfigwidth}
    \includegraphics[width=\linewidth]{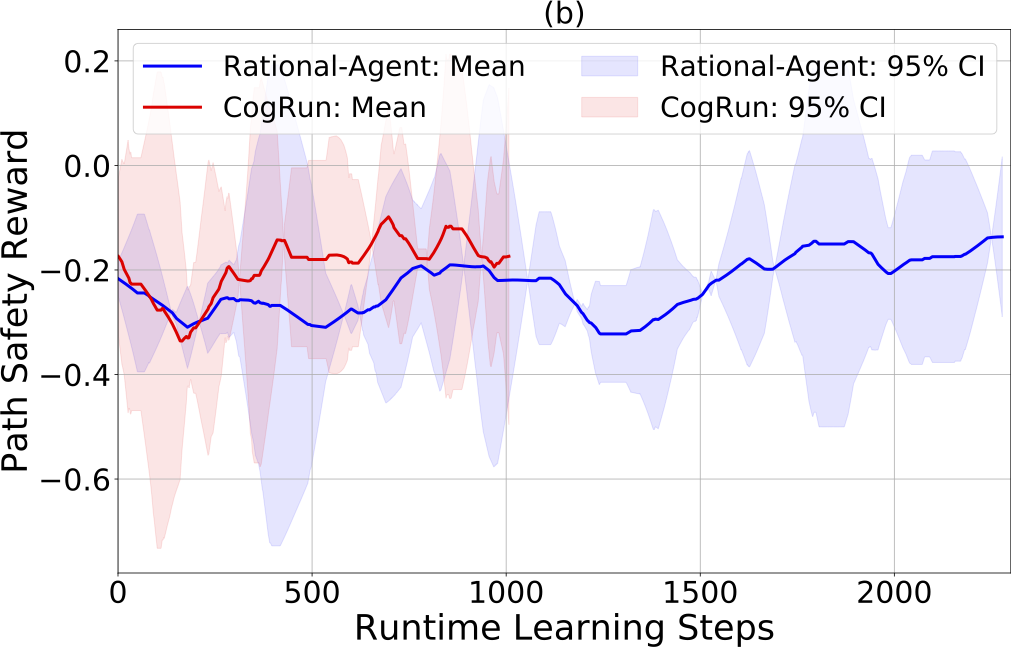}
\end{subfigure}
\hfill
\begin{subfigure}{\subfigwidth}
\includegraphics[width=\linewidth]{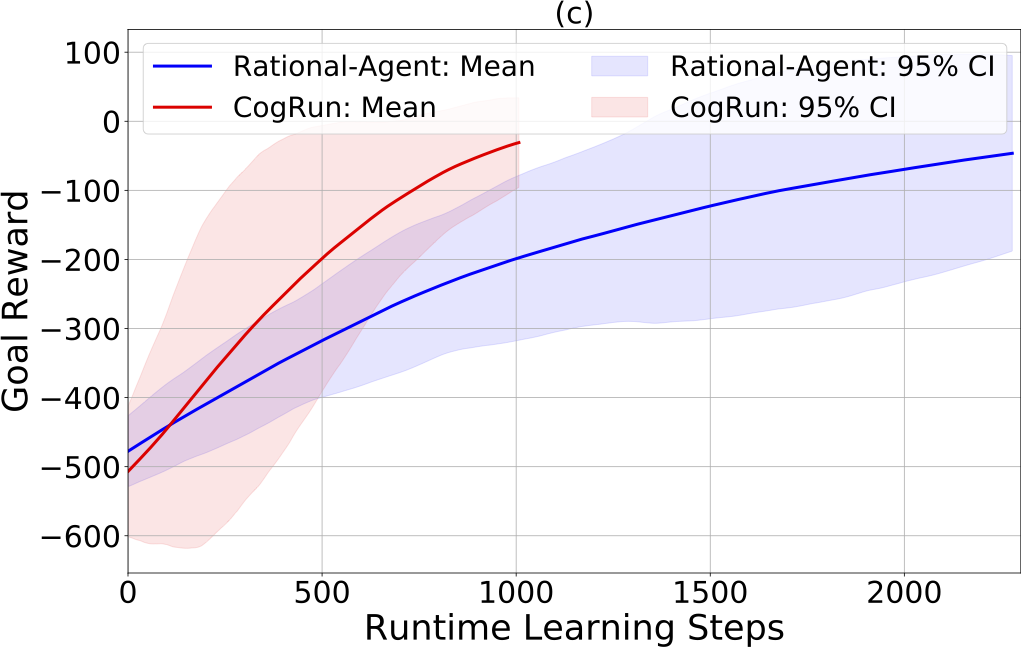}
\end{subfigure}
\hfill
\begin{subfigure}{\subfigwidth}
    \includegraphics[width=\linewidth]{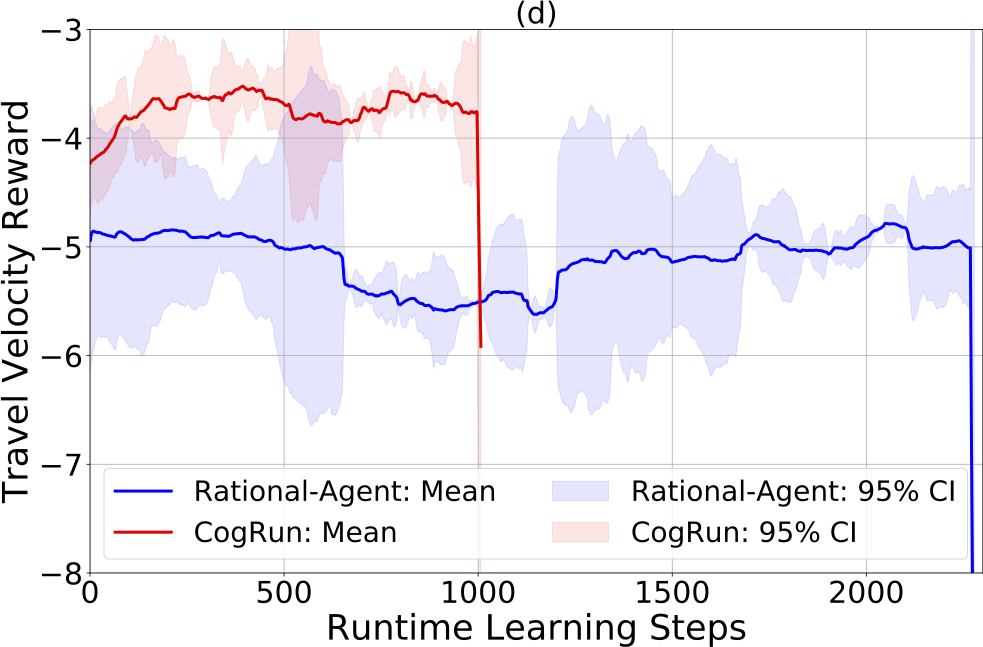}
\end{subfigure}
\vspace{-0.1cm}
\caption{Performance metics for body stability, path safety, goal reaching, and travel velocity.}
\label{meanreward}
\vspace{-0.3cm}
\end{figure}
We recall that the Rational-Agent is a non-learning module, and its capability remains fixed after deployment. The step-wise rewards in Fig. \ref{meanreward} shows that \textbf{CogRun} progressively achieves smoother trajectories, higher rewards, and faster goal-reaching than the Rational-Agent, validating its effectiveness for safe runtime adaptation.

\subsubsection{\textbf{Runtime Learning for End-to-End Vision-to-Action}} In this task, the safety set is the same as that used in the previous task. The reward is $r = \xi_1 \cdot \mathcal{R}_{bs} + \xi_2 \cdot \mathcal{R}_{ps} + \xi_3 \cdot \mathcal{R}_{gr} +  \xi_4 \cdot \mathcal{R}_{RA} + \xi_5 \cdot \mathcal{R}_{vel} + \xi_6 \cdot \mathcal{R}_{jerk}$, where the $\mathcal{R}_{jerk} = -\left\| \mathbf{a}_{\text{learning}}(k) - \mathbf{a}_{\text{learning}}(k-1) \right\|_2^2$ is newly included for encouraging smooth motions by penalizing abrupt changes in consecutive actions. Its associated weight $\xi_6 = 0.0002$. Other reward components and coupled weights are the same as the ones in the previous section.  The input of Learning-Agent is the same as the one in previous task, but its output is a 6-DoF acceleration command $\mathbf{a}=[\dot{\mathbf{v}},\dot{\boldsymbol{\omega}}]$. The Rational Agent consists of Interactive-FAR~\cite{he2024interactive} and physics-based adaptive controller in \cite{realdrl}. 

\begin{figure}[H]
\vspace{-0.152cm}
\centering
\includegraphics[width=0.49\textwidth]{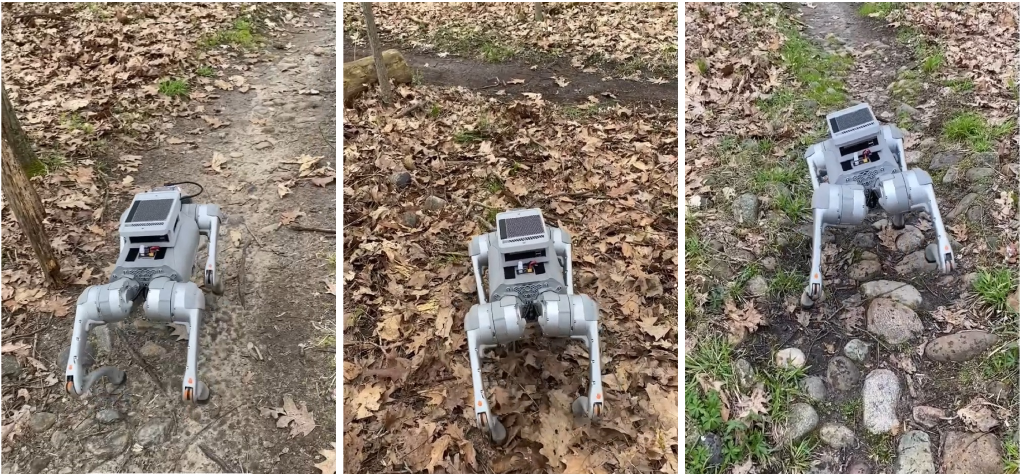}
\caption{Three different physical environments for locomotion learning.
\textbf{Left}: A forest path with a mix of rounded stones and packed dirt.
\textbf{Middle}: A densely leaf-covered forest floor with scattered branches, twigs, and stones.
\textbf{Right}: A narrow, rocky path with rounded and dense cobblestones.}
\label{env2}
\vspace{-0.148cm}
\end{figure}

This section evaluates whether \textbf{CogRun} improves early-stage safety and maintain safe runtime learning within both unknown and dynamic environments. We first compare \textbf{CogRun} with three fault-tolerant RL frameworks, namely Real-DRL \cite{realdrl}, Runtime Learning Machine (RLM) \cite{runtimemao}, and Model Predictive Shielding (MPS) \cite{bastani2021safe,banerjee2024dynamic}, in the same unknown environment. The experimental physical environment is shown in Fig. \ref{env2} (left), consisting of a forest path with randomly distributed rounded stones and packed dirt. 

The robot behaviors across three runtime learning episodes of 10, 15, and 20 are available via \href{https://www.dropbox.com/scl/fi/pyf6gznuaz9aeij0y08m2/end2end.mp4?rlkey=n962dly2dc04ja6mnpuchbgis&st=8zopxdhb&dl=0}{\color{blue}[vision-to-action-learning-comparison link]}. These results highlight the critical importance of
safe runtime learning for safety-critical robots: safety violations (including loss of balance and collisions) can terminate the learning process. Meanwhile, the quantitative performance of runtime learning over 20 consecutive trips or episodes is summarized in Table \ref{table:exp_result}. These results show that \textbf{CogRun} substantially improves safety and learning stability in unknown physical environments, compared with existing state-of-the-art fault-tolerant RL frameworks.

\begin{table}[http]
\label{table:exp_result}
\centering
\resizebox{0.98\columnwidth}{!}{
\begin{tabular}{c|cccc}
\toprule
\textbf{Model ID} & \textbf{Lose Balance} & \textbf{Collision} & \textbf{Dead-Zone Encounter} & \textbf{Task Completion} \\\midrule
\textbf{CogRun} & \textcolor{red}{0/20} & \textcolor{red}{0/20} & \textcolor{red}{1/20} & \textcolor{red}{19/20}     \\  \midrule
\textbf{Real-DRL} & 7/20 & 2/20 & 1/20 & 10/20 \\  \midrule
\textbf{MPS} & 6/20 & 4/20 & 0/20 & 10/20 \\  \midrule
\textbf{RLM} & 6/20 & 3/20 & 0/20 & 11ß/20 \\  \bottomrule
\end{tabular}}
\vspace{-0.cm}
\caption{Statistics of runtime learning for vision-to-action policy over 20 trips or episodes of runtime learning. Note:  dead zones are physically reachable areas whose environmental demands exceed a robot's capabilities for safe task execution, and ``\textit{Task Completion = No Collision + No Loss of Balance + No Dead-Zone Encounter + Goal Reached.}" } \label{table:exp_result}
\vspace{-0.25cm}
\end{table}

Finally, we also evaluate \textbf{CogRun}'s robustness to non-stationary or dynamic environments. Specifically, the robot launches runtime learning across three distinct environments shown in Fig. \ref{env2}. Specifically, after five episodes in Environment~1, the learned policy is directly transferred to Environments~2 and~3 without reinitialization. The robot's learning behavior is available via \href{https://www.dropbox.com/scl/fi/5k0xvijs7qb0hp2zdt2yy/env.mp4?rlkey=94w7hxu034uien526mzqd7jxz&st=7bw6pefd&dl=0}{\color{blue}[different-environments-learning link]}, indicating \textbf{CogRun} consistently maintains safe runtime learning across diverse non-stationary and unknown physical environments.

\subsection{Autonomous Vehicles in Simulated Unknown Wild Forests}
This Gazebo simulation on an off-road autonomous vehicle has two objectives: i) demonstrate \textbf{CogRun}'s generalizability across different robotic platforms, and ii) evaluate the contributions of its key cognitive mechanisms through an ablation study. The simulation runs on a high-end PC (Intel Core i9 CPU, NVIDIA RTX PRO 6000 Blackwell GPU). The RL configuration is shown in Table \ref{table:specs}.

\begin{figure}[H]
\vspace{-0.2cm}
\centering
\includegraphics[width=0.48\textwidth]{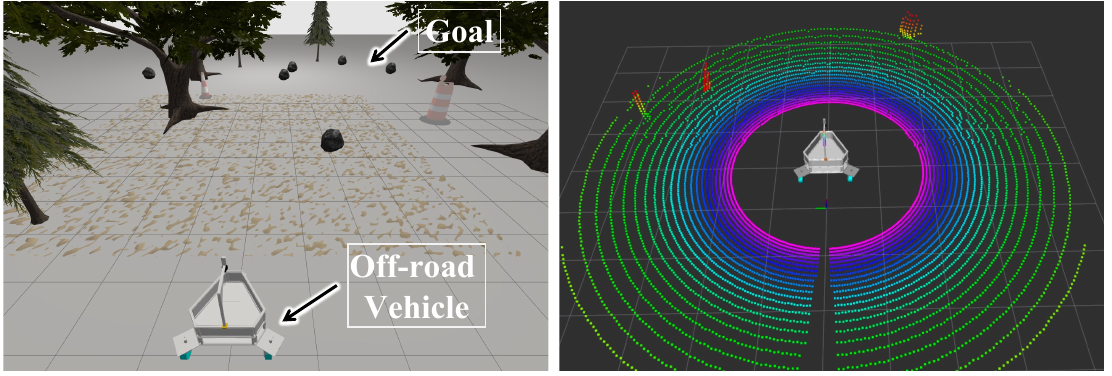}
\caption{Simulation setup in Gazebo (Left) and Rviz (Right).}
\label{simulationsetup}
\vspace{-0.15cm}
\end{figure}

As shown in Figure \ref{simulationsetup}, the off-road vehicle is a three-wheel swerve-drive robot navigating through randomized unknown wild-forest environments containing rocks, trees, and terrain ranging from flat to uneven. For the ablation study, we consider three configurations: (1) cognition-driven batch sampling only (denoted as `\ul{Cognition}'); (2) cognition-driven batch sampling together with safety-aware RL-IBL blending (denoted as `{Cognition+Blending}'); and (3) neither mechanism, which reduces `\textbf{CogRun}' to conventional fault-tolerant RL with random batch sampling (denoted as `{Random}'). Note: these is no `\ul{Random+Blending}', because '{Blending}' builds on `{Cognition}', i.e., the `\ul{Blending}' if unavailable is there is no `{Cognition}'. 

The runtime learning task is for navigation. The safety set is defined as:
$\mathbb{S}=\{\mathbf{s}\mid d_{min}\ge0.55~\mathrm{m},~
|\dot{\theta}|\le1~\mathrm{rad/s},
|\dot{\phi}|\le1~\mathrm{rad/s}\}$, where $\dot{\theta}$ and $\dot{\phi}$ are the pitch and roll rate, respectively.
The Rational-Agent adopts Interactive-FAR \cite{he2024interactive}, while the Learning-Agent acquires a safe and efficient navigation policy through runtime learning with the reward as $r = \xi_1 \cdot \mathcal{R}_{bs} + \xi_2 \cdot \mathcal{R}_{ps} + \xi_3 \cdot \mathcal{R}_{pl} + \xi_4 \cdot \mathcal{R}_{RA} + \xi_5 \cdot \mathcal{R}_{vel}$,
where $\mathcal{R}_{bs}$, $\mathcal{R}_{ps}$, $\mathcal{R}_{pl}$, $\mathcal{R}_{RA}$, and $\mathcal{R}_{vel}$ are defined in Eqs. \eqref{sta}--\eqref{velocity}, respectively. Their associated parameters are $\xi_1 = 0.1$, $\xi_2 = 0.05$, $\xi_3 = 0.01$, $\xi_4 = 0.3$, $\xi_5 = 0.002$. 

\begin{figure}[H]
\vspace{-0.3cm}
    \centering
    \includegraphics[width=0.95\columnwidth]{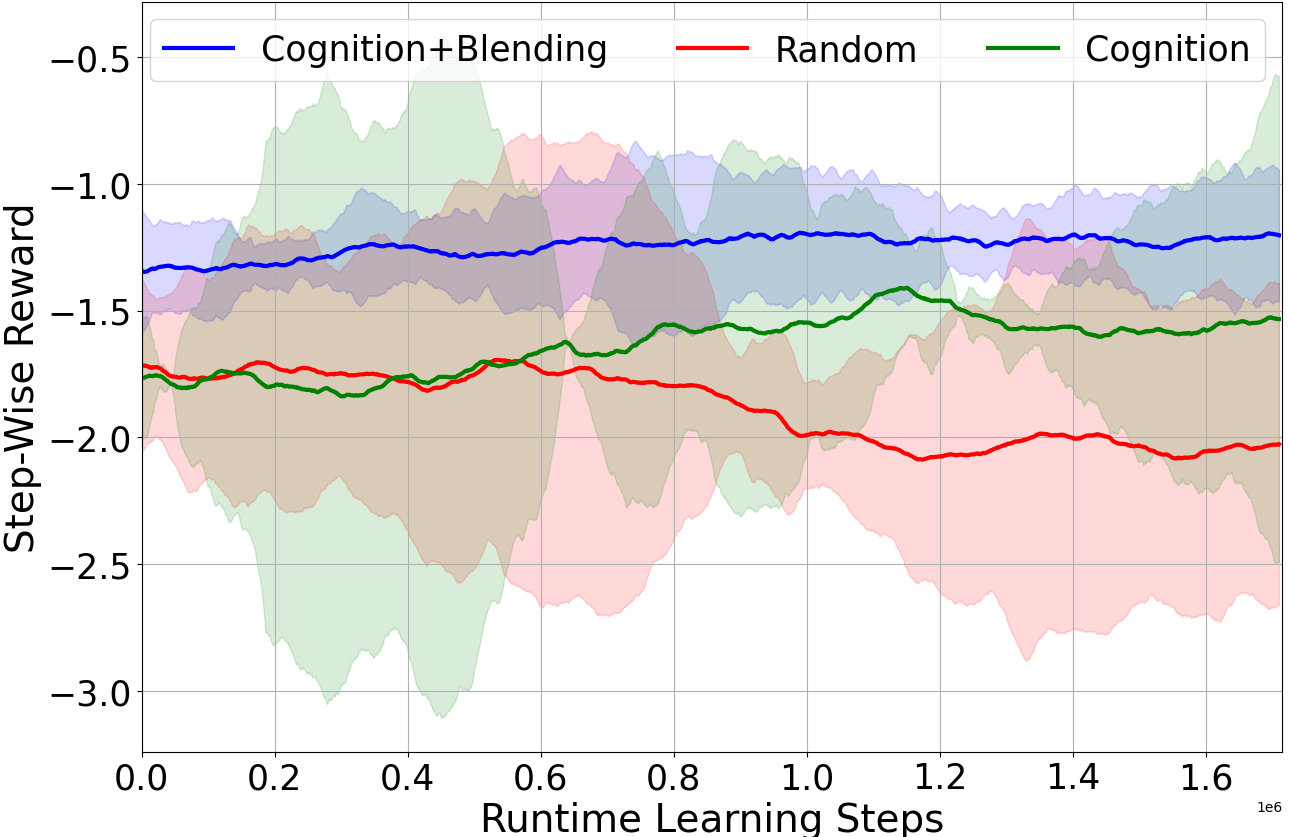}
    \caption{The stepwise rewards (95\% CI) of the off-road vehicle over five random seeds, each with a total of 60 episodes.}
    \label{finalaamain}
    \vspace{-0.15cm}
\end{figure}

The ablation results in Fig. \ref{finalaamain} indicate that, `\ul{Cognition}' consistently outperforms `\ul{Random}', while `\ul{Cognition+Blending}' further achieves the highest rewards. Robot behaviors across three consecutive runtime learning episodes are available via \href{https://www.dropbox.com/scl/fi/noqvo6f3iuj34r61g5f5a/acomcar.mp4?rlkey=jgzgwa6amvog7jkivsvcvkyz9&st=v1entonq&dl=0}{\color{blue}[vehicle-learning-nav link]}, demonstrating that the proposed safety-aware RL-IBL blending mechanism (building on the cognition-driven batch sampling) yields significantly safer and higher-performance runtime learning. 

\section{Conclusion and Discussion} \label{cdfin}
This paper presents \textbf{CogRun}, a cognitively-grounded on-device runtime learning framework for ground robots operating in unknown physical environments. Its key innovations include a cognitive-neural learning architecture with dedicated replay buffers, cognition-driven batch sampling, and safety-aware RL-IBL blending, enabling safer, more efficient, and fully on-device runtime learning. Real-world experiments on a quadruped robot and simulation studies on an off-road autonomous vehicle demonstrate the effectiveness and generalizability of the proposed framework. Moving forward, we will investigate principled experience ranking and memory decay mechanisms, as well as more effective measures of frequency, recency, and similarity, to further advance cognitively grounded runtime learning for physical AI robotic systems.

\bibliographystyle{unsrt}
\bibliography{reference}

\end{document}